\documentclass[runningheads]{llncs}
\usepackage{graphicx}
\usepackage{amsmath,amssymb} 
\DeclareMathOperator*{\argmin}{arg\,min}
\usepackage{color}
\usepackage{ulem}
\usepackage[width=122mm,left=12mm,paperwidth=146mm,height=193mm,top=12mm,paperheight=217mm]{geometry}
\begin{document}
\pagestyle{headings}
\mainmatter

\title{Head Reconstruction from Internet Photos} 

\titlerunning{Head Reconstruction from Internet Photos}

\authorrunning{Shu Liang, Linda G. Shapiro, Ira Kemelmacher-Shlizerman}

\author{Shu Liang, Linda G. Shapiro, Ira Kemelmacher-Shlizerman}


\institute{Computer Science \& Engineering Department,\\
        University of Washington\\
        \email{ \{liangshu,shapiro,kemelmi\}@cs.washington.edu }
}

\maketitle

\begin{abstract}
3D face reconstruction from  Internet photos  has recently produced exciting results. A person's face, e.g., Tom Hanks, can be modeled and animated in 3D from a completely uncalibrated  photo collection. Most methods, however, focus solely on face area and mask out the rest of the head. This paper proposes that head modeling from the Internet is a problem we can solve. We target reconstruction of the rough shape of the head. Our method is to  gradually ``grow'' the head mesh starting from the frontal face and extending to the rest of views using photometric stereo constraints. We call our method boundary-value growing algorithm. Results on photos of celebrities downloaded from the Internet are presented.
\end{abstract}

\section{Introduction}

\begin{quotation}
	``If two heads are better than one, then what about double chins? On that note, I will help myself to seconds.'' ---Jarod Kintz
\end{quotation}

Methods that reconstruct  3D models of people's heads from images need to account for varying 3D pose, lighting, non-rigid changes due to expressions, relatively smooth surfaces of faces, ears and neck, and finally, the hair. Great reconstructions can be achieved nowadays in case the input photos are captured in a calibrated lab setting or semi-calibrated setup where the person has to participate in the capturing session (see related work). Reconstructing from Internet photos, however, is  an open problem due to the high degree of variability across  uncalibrated photos; lighting, pose, cameras and resolution change dramatically across photos.  In recent years, reconstruction of \textit{faces} from the Internet have received a lot of attention.  All face-focused methods, however, mask out the head using a fixed face mask and focus only on the face area. For real-life applications, we must be able to reconstruct a full head.  
 
So what is it there to reconstruct except for the face? At the minimum, to create full head models we need to be able to reconstruct the ears, and at least part of the neck, chin, and overall head shape.  Additionally, hair reconstruction is a difficult problem. One approach is to use morphable model methods. These, however, do not fit the head explicitly but instead use fitting based on the face and provide a mostly average (non-personalized) bald model for the head. 

This paper addresses the new direction of \textit{head} reconstruction directly from Internet data. We propose an algorithm to create a rough head shape, and frame the problem as follows. Given a photo collection, obtained by searching for photos of a specific person on Google image search, we would like to reconstruct a 3D model of that person's head. Just like \cite{kemelmacher2011face} (that focused only on the face area) we aim to reconstruct an average rigid model of the person from the whole collection. This model can be then used as a template for dynamic reconstruction, e.g., \cite{suwajanakorn2014total} , and hair growing techniques, e.g., \cite{hu2015single}. Availability of a template model is  essential  for those techniques. 
 
 \begin{figure}[t]	
\includegraphics[width=\columnwidth]{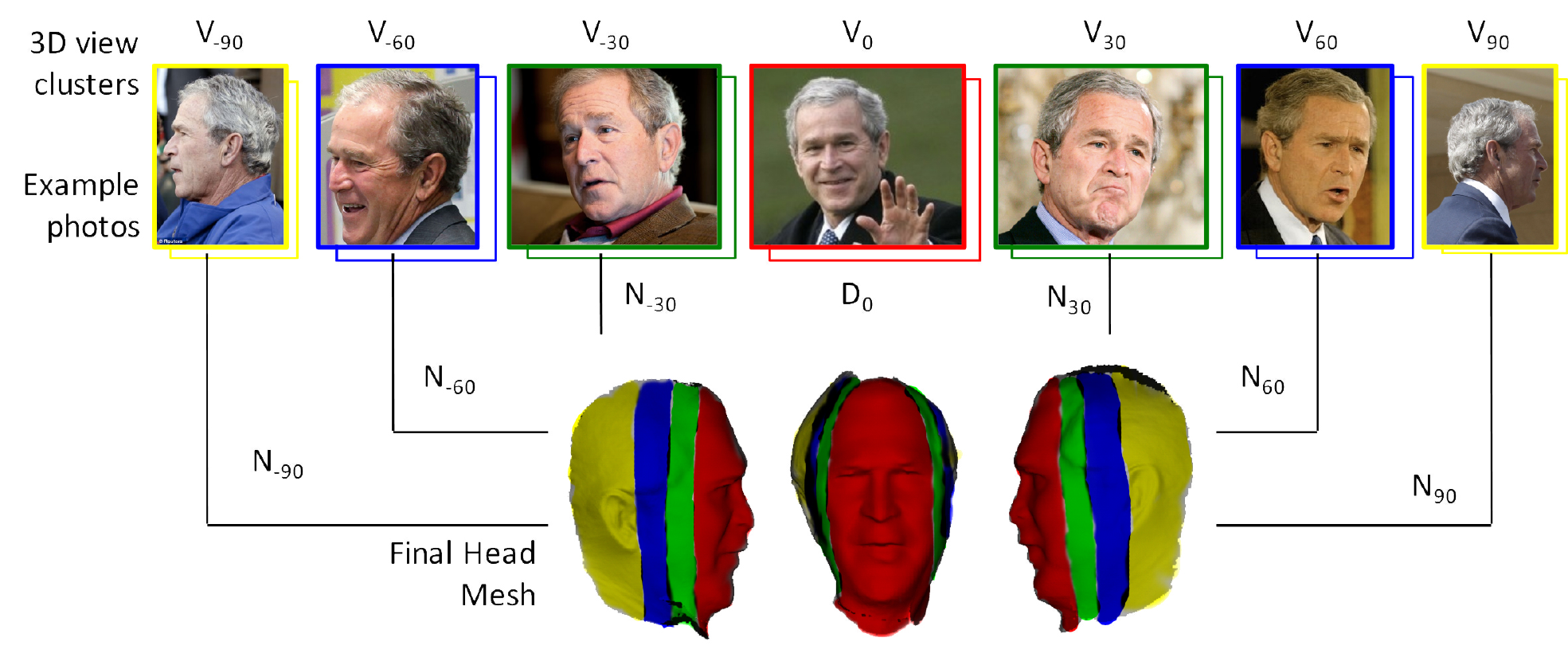}
 	\caption{By looking at the top row photos we can imagine how Bush's head shape looks in 3D; however, existing methods fail to do so on Internet photos, due to such facts as inconsistency of lighting, challenging segmentation,and expression variation. Given many more photos per 3D view (hundreds), however, we show that a rough full head model can be reconstructed. The head mesh is divided into 7 parts, where each part is reconstructed  from a different view cluster while being constrained by the neighboring view clusters.}\label{fig:system}
 \end{figure}
 
Consider the top row  photos in Fig.~\ref{fig:system}. The 3D shape of the head is clearly outlined in the different views (different 3D poses).  However, if we are given only one or two photos per view, the problem is still very challenging due to lighting inconsistency across views, difficulty in segmenting the face profile from the background, and challenges in merging the images across views. Our key idea is that with many more (hundreds) of photos per 3D view, the challenges can be overcome.  For celebrities, we can easily acquire such collections from the Internet; for others, we can extract such photos from Facebook or from mobile photos.

Our method works as follows: A person's photo collection is divided to clusters of approximately the same azimuth angle of the 3D pose.  Given the clusters,  a depth map of  the frontal face is reconstructed, and the method  gradually grows the  reconstruction by estimating surface normals per view cluster and then constraining using boundary conditions coming from  neighboring views. The final result is a head mesh of the person that combines all the views.

\section{Related Work}

The related work is in calibrated and semi-calibrated setting for head reconstruction, and uncalibrated settings for face reconstruction.  

Calibrated head modeling has achieved amazing results over the last decade \cite{debevec2012light,alexander2013digital,beeler2010high}. Calibrated methods require a person to participate in a capturing session to achieve good results. These typically take as input a video  with relatively constant lighting, and  large pose variation across the video.  Examples include non rigid structure from motion methods \cite{agudo2014online,garg2013dense}, multiview methods \cite{tanskanen2013live,ichim2015dynamic}, dynamic kinect fusion \cite{newcombe2015dynamicfusion},  and RGB-D based methods \cite{thies2015real,zollhofer2014real}. 

Reconstruction of people from Internet photos recently achieved good results; \cite{kemelmacher20113d} showed that it is possible to reconstruct a face from a single Internet photo using a template model of a different person.  \cite{kemelmacher2011face}  later proposed a photometric stereo method to reconstruct a face from many Internet photos of the same individual. \cite{roth2015unconstrained} showed that photometric stereo can be combined with face landmark constraints,  and recent work has shown that  3D dynamic shape \cite{suwajanakorn2014total,garrido2013reconstructing,shi2014automatic} and texture \cite{suwajanakorn2015makes} can be recovered from Internet photos. 

One way to approach the uncalibrated head reconstruction problem  is to use the morphable model approach. With morphable models 
\cite{blanz1999morphable,hsieh2015unconstrained},  the face is fitted to a linear space of 200 face scans, and the head is reconstructed  from the linear space as well.  In practice, morphable model methods work well for face tracking \cite{shapiro2014rapid,li2013realtime}. However, there is no actual fitting of the head, ears, and neck of the person to the model, but rather an approximation derived from the face; thus the reconstructed model is not personalized. A morphable model for ears \cite{bustard20103d} was proposed, but it was not applied to uncalibrated Internet photos.

Hair modeling  requires a multiview calibrated setup \cite{luo2013structure,hu2014robust} or can be done from a single photo by fitting to a database of synthetic hairs \cite{hu2015single}, or by fitting helices \cite{chai2015high,chai2012single}. Hair reconstruction methods assume that the user marks hair strokes or that a rough model of the head, ears and face is provided. The goal of this paper is to provide such a rough head shape model; thus our method is complementary to hair modeling techniques.

\section{Overview}

\begin{figure}
	\includegraphics[width=\textwidth]{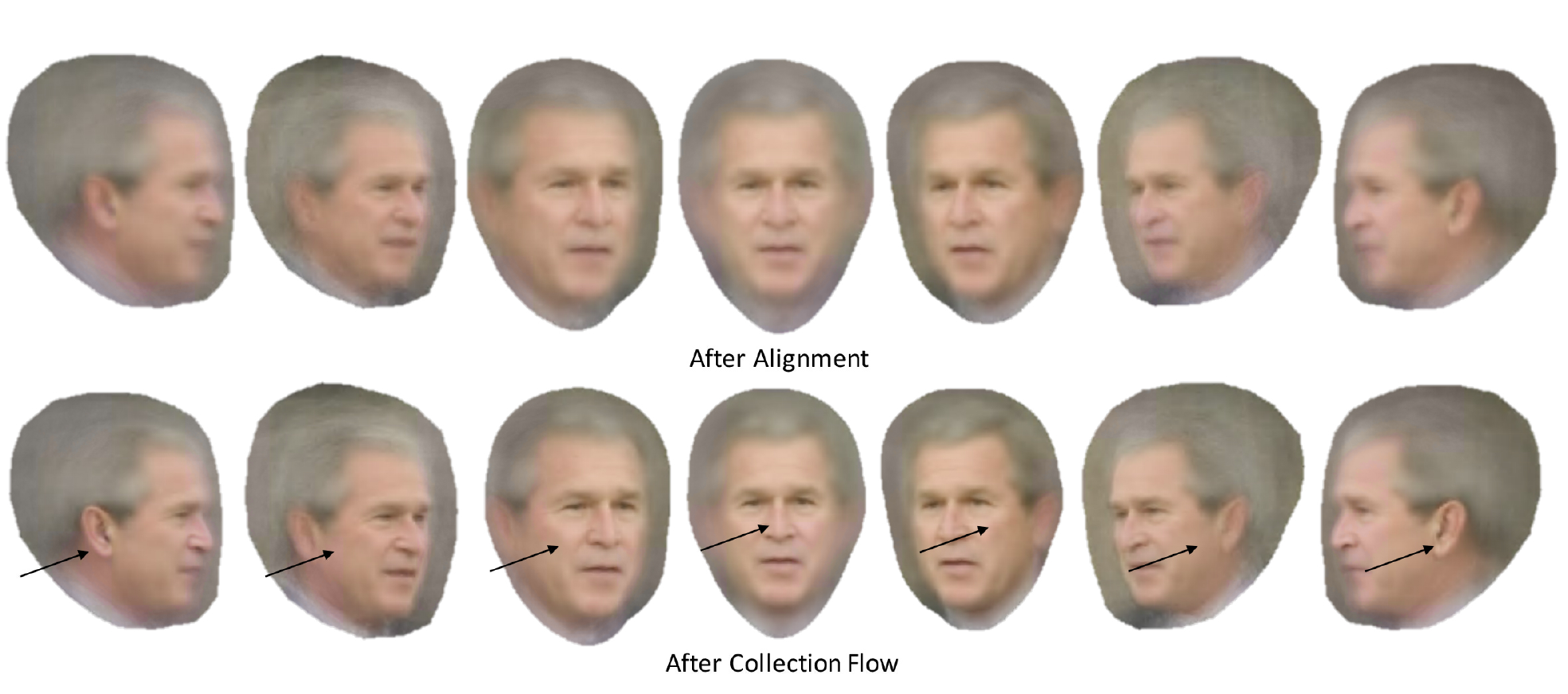}
	\caption{Averages of view clusters' photos after rigid alignment (1st row) and after collection flow (2nd row). The arrows visualize  head parts  that are sharper in each view, e.g., the ear is sharpest in 90 and -90 degrees (left and right). The key idea is to use the sharp (well-aligned) parts from the corresponding views to create an optimal mesh reconstruction.}\label{fig:averages}
\end{figure}

We denote the set of photos in a  view cluster as $V_i$. Photos in the same view cluster have approximately the same 3D pose and azimuth angle. Specifically, we divided the photos into $7$ clusters with azimuths: $i={0,-30,30,-60,60,-90,90}$.  Figure~\ref{fig:averages} shows the averages of each cluster after rigid alignment using fiducial points (1st row) and after subsequent alignment using the Collection Flow method \cite{kemelmacher2012collection} (2nd row), which calculates optical flow for each cluster photo to the cluster average.  A key observation is that each view cluster has one particularly well-reconstructed head area, e.g., the ears in views 90 and -90  are sharp while blurry  in other views.   Since our goal is to create a full head mesh, our algorithm will combine the optimal parts from each view into a single model. This is illustrated in Figure~\ref{fig:system}.  

It was shown in previous work that the face can be reconstructed from frontal photos using Photometric Stereo \cite{kemelmacher2011face}. Thus, one way to implement our idea, of combining views into a single mesh, would be to reconstruct shape from each view cluster independently and then stitch them together. This turned out to be challenging as the individual shapes are reconstructed up to linear ambiguities. Although the photos are divided into pose clusters, the precise pose for each pose cluster is unknown. For example, $V_{30}$ could have a variance from 25 to 35 in the azimuth rotation angle, depending on the dominant pose of the image cluster. This misalignment will also increase the difficulty of stitching all the views. We solve those challenges by growing the shape in stages works well.  We begin by describing estimation of surface normals and a depth map for view cluster $V_0$ (frontal view) in section~\ref{sec:initial}.  This  will be the initialization for our algorithm. In section~\ref{sec:alg}, we describe how each view cluster uses its own photos and the depth of its neighbors to contribute to the creation of a full head mesh.  Data acquisition and alignment details are given in the experiments section (Section~\ref{sec:results}).

\section{Head Mesh Initialization}\label{sec:initial}

Our goal is to reconstruct the head mesh $M$. We begin  by estimating a depth map and surface normals of the frontal cluster $V_0$, and assign each reconstructed pixel to a vertex of the mesh.  The depth map is estimated by extending the method of \cite{kemelmacher2011face}  to capture more of the head in the frontal face photos, i.e. , we extend the reconstruction mask to a bigger area to capture the chin, part of the neck and some of the hair. The algorithm is as follows: 
\begin{enumerate}
\item \textbf{Dense 2D alignment:} Photos are first rigidly aligned using 2D fiducial points as the pipeline of \cite{kemelmacher2011exploring}. The head region including neck and shoulder in each image is segmented using semantic segmentation by \cite{zheng2015conditional}. Then Collection Flow \cite{kemelmacher2012collection} is run on all the photos in $V_0$ to densely align them to the average photo of that set. Note that the segementation works remarkably well on most photos. The challenging photos do not affect our method; given that the majority of the photos are segmented well, Collection Flow will correct for inconsistencies. Also, Collection Flow helps overcome differences in hair style by warping all the photos to the dominant style. See more details about alignment in Section~\ref{sec:results}.\\

	\item \textbf{Surface normals estimation:} We used a template face mask to find the face region on all the photos. Photometric Stereo (PS) is then applied to the face region of the flow-aligned photos. The face region of the photos are arranged in an $n \times p_k$ matrix $Q$, where $n$ is the number of images and $p_k$ is the number of face pixels determined by the template facial mask. Rank-4 PCA is computed to factorize into lighting and normals: $Q = LN$. After we get the lighting estimation $L$ for each photo, we can compute N for all $p$ head pixels including ear, chin and hair regions.\\

Two key components that made PS work on uncalibrated head photos are:\\

1) resolving the Generalized Bas-Relief (GBR) ambiguity using a template 3D face of a different individual, i.e., $\min_A ||N_{\text{template}}-AN_{face}||^2$, \\

2) using a per-pixel surface normal estimation, where each point uses a different subset of photos to estimate the normal. We follow the per-pixel surface estimation idea as in previous work, i.e., given the initial lighting estimate $L$, the normal is computed per point by selecting a subset of $Q$'s rows that satisfy the re-projection constraint. In the full head case, we extend it to handle cases when the head is partially cropped out, by adding a constraint that a photo participates in normal estimation if it satisfies both the reprojection constraint and is inside the desired head area, i.e., part of the segmentation result from \cite{zheng2015conditional}. If the number of selected subset images is not enough (less than $n/3$), we will not use them in our depth map estimation step.\\

	\item \textbf{Depth map estimation:} The surface normals are integrated to create a depth map $D_0$ by solving a linear system of equations that satisfy gradient constrains  $dz/dx  = -n_x/n_y$ and $dz/dy = -n_x/n_y$ where ($n_x,n_y,n_z$) are components of the surface normal of each point \cite{basri2007photometric}. Combining these constraints, for the $z$-value on the depth map, we have:
\begin{equation}
n_z(z_{x+1,y}-z_{x,y})=n_x
\end{equation}
\begin{equation}
n_z(z_{x,y+1}-z_{x,y})=n_y
\end{equation}
In the case of $n_z \approx 0$, we use a different constraint,
\begin{equation}
n_y(z_{x,y}-z_{x+1,y})=n_x(z_{x,y}-z_{x,y+1}) 
\end{equation}
This generate a sparse matrix of $2p \times 2p $ matrix M, and we can solve for:
\begin{equation}
\argmin_{z} {||Mz-v||}^2
\end{equation}
We do a least squares fit to solve for the $z$-value for each pixel.

\end{enumerate}
Potentially, we could run the same algorithm for each view cluster. This, however, does not perform well, as we will see in the experiments section. Instead we are going to introduce two constraints, which we describe in the next section. 

\section{Boundary-Value Growing}\label{sec:alg}
In this section we describe our ``growing'' algorithm to complete the side views of the mesh. Starting from the frontal view mesh $V_0$, we gradually complete more regions of the head in the order of $V_{30}$, $V_{60}$, $V_{90}$ and $V_{-30}$, $V_{-60}$, $V_{-90}$. For each view cluster we repeat the same algorithm as in Section~\ref{sec:initial} with two additional key constraints: 

\begin{enumerate}
	
	\item  \textbf{Ambiguity recovery:} Rather than recovering the ambiguity $A$ that arises from $Q=LA^{-1}AN$ using the template model, we use the already computed neighboring cluster, i.e., for $V_{\pm 30}$, $N_0$ is used, for $V_{\pm 60}$ we use $N_{\pm 30}$, and for $V_{\pm 90}$ we use $N_{\pm 60}$.  Specifically, we estimate the out-of-plane pose from our 3D initial mesh $V_0$ to the average image of pose cluster $V_{30}$ using the method proposed in \cite{suwajanakorn2014total}. We render the rotated mesh $V_0'$ as a reference depth map $D_0'$ to pose cluster $V_{30}$, accounting for visibility and occlusion using zbuffer. The normals on each projected pixels of $D_0'$ will serve as the reference normals to solve for the GBR ambiguity of the overlapping head region as well as the newly grown head region. \\

	\item \textbf{Depth constraint:} In addition to the gradient constraints that are specified in Sec.~\ref{sec:initial}, we modify the boundary constraints from Neumann to Dirichlet. Let $\Omega_{0}$ be the boundary of $D_0'$. Then we impose that the part of $\Omega_{0}$ that intersects the mask of  $D_{30}$ will have the same depth values: $D_{30}(\Omega_{0})=D_{0}'(\Omega_0)$. With both boundary constraints and gradient constraints, our optimization function can be written as:
\begin{equation}
\argmin_{z}{{||Mz-v||}^2+{||Wz-Wz_0||}^2}
\end{equation}

where $z_0$ is the depth constraint from $D_{0}'$, and $W$ is a blend mask with values decreasing from $1$ to $0$ on the boundary of $D_{0}'$. We will get the new vertex positions for grown regions and we can also update vertices on the boundary of the already computed depth map, eliminating the distortion caused by lack of photos and inaccurate $n_z$. This process is repeated for every neighboring pair of depths. 
\end{enumerate}

After each depth stage reconstruction (0,30,60,.. degrees),  the estimated depth is projected to the head mesh. By this process, the head is gradually filled in by gathering vertices from all the views.

\section{Experiments}\label{sec:results}

We describe the data collection process, alignment, evaluations and comparisons with other methods. 
 
\subsection{Data Collection and Processing}
We collected around $1,000$ photos per person (George Bush, Vladimir Putin, Barack Obama and Hillary Clinton) by searching for photos on Google image search. The numbers of images in each pose cluster are shown in Table \ref{table:Nphotos}. We noticed that the numbers of side view photos are usually much smaller than frontal view photos. In order to get more photos, we searched for ``Bush shakes hands'', ``Bush shaking hand'', ``Bush portrait'', ``Bush meets'' etc. to collect more non-frontal photos. The number of photos in each cluster will affect the final result; we will demonstrate the reconstruction quality vs. number of photos later in this section.

\begin{table}
\begin{center}
\caption{Number of photos we used in each pose cluster}
\label{table:Nphotos}
\begin{tabular}{llllllll}
\hline\noalign{\smallskip}
Pose &-90 &-60 &-30 &0 &30 &60 &90\\
\noalign{\smallskip}
\hline
\noalign{\smallskip}
Bush &185 &62 &118 &371 &113 &80 &191\\
Putin &131 &58 &151 &413 &121 &61 &151\\
Obama &65 &51 & 126 &284 &177 &55 &75\\
Clinton &115 &47 &114 &332 & 109 &61 &66\\
\hline
\end{tabular}
\end{center}
\end{table}

We ran face detection and fiducial detection using IntraFace\cite{xiong2013supervised}. For extreme side views, none of the state of the art fiducial detection algorithms was able to perform, and often times the face was not even detected. We therefore manually annotated each photo with 7 fiducials. 

Once photos are aligned we run collection flow \cite{kemelmacher2012collection} on each view cluster. For completeness we review the method. The idea is  to estimate a lighting subspace from all the photos in a particular cluster $V_i$ via PCA. Then each photo in the cluster $V_i^j$ is projected to the subspace to produce photo $\hat{V_i^j}$, which has a similar lighting as $V_i^j$ but an average shape.  Optical flow is then estimated between $V_0^j$ and its relighted version $\hat{V_0^j}$. The process is iterated over the whole collection.  In the end, all photos are warped to approximately average shape; however, they retain their lighting which makes them amenable for photometric stereo methods. \\
 
 \subsection{Results and Evaluation}
\begin{figure}
 	\includegraphics[width=\textwidth]{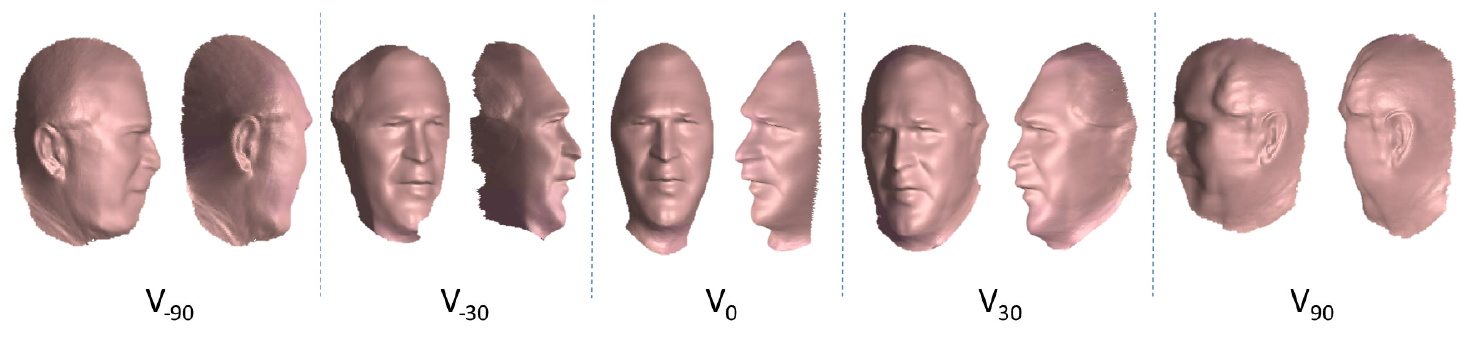}
 	\caption{Individual reconstructions per view cluster, with depth and ambiguity constraints. We can see that the individual views provide different shape components. For each view we show the mesh in two poses.}\label{fig:view3d}
 \end{figure}

 Fig.~\ref{fig:view3d} shows the reconstruction per view that was later combined to a single mesh. For example, the ear in 90 and -90 views is reconstructed well, while the other views are not able to reconstruct the ear. 

\begin{figure}
 	\includegraphics[width=\textwidth]{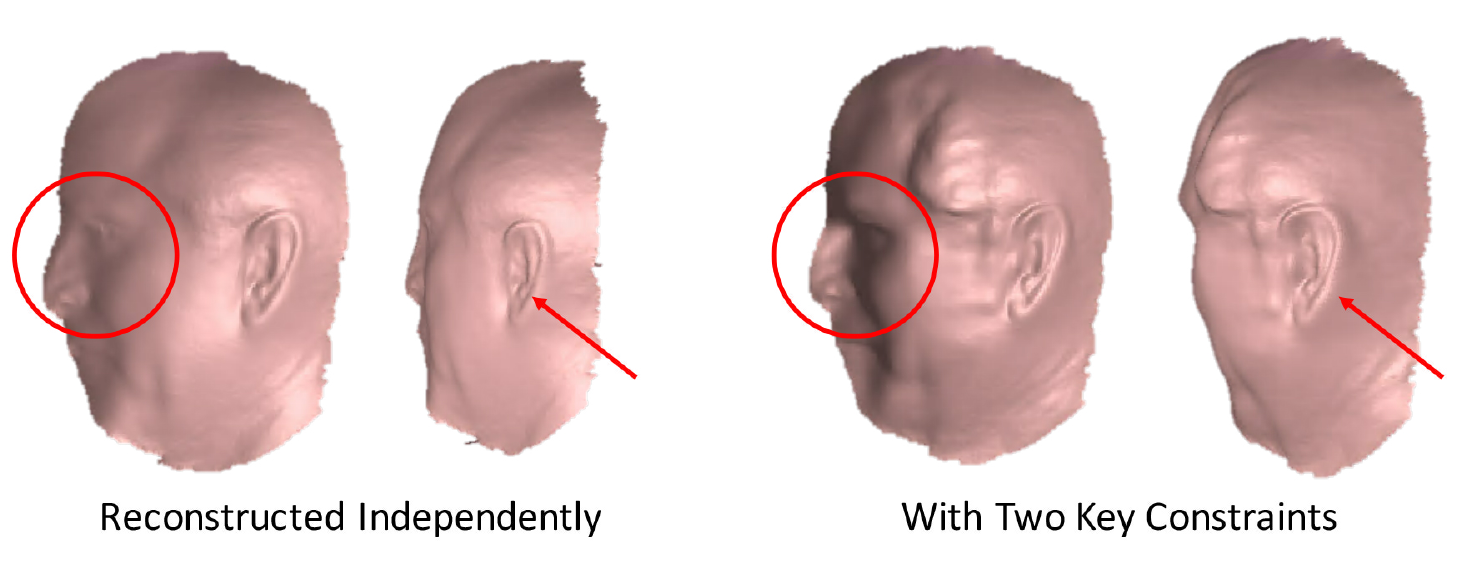}
 	\caption{Comparison between without and with two key constraints. The left two shapes show the two views of 90 degree view shape reconstructed independently without two key constraints. The right two shapes show the two views of our result with two key constraints. } \label{fig:constraints}
 \end{figure}

In Figure~\ref{fig:constraints}, we shows how our two key constraints work well in the degree 90 view reconstruction result. Without the correct reference normals and depth constraint, the reconstructed shape is flat and the profile facial region is blurred, which increased the difficulty of aligning it back to the frontal view.

\begin{figure}
 	\includegraphics[width=\textwidth]{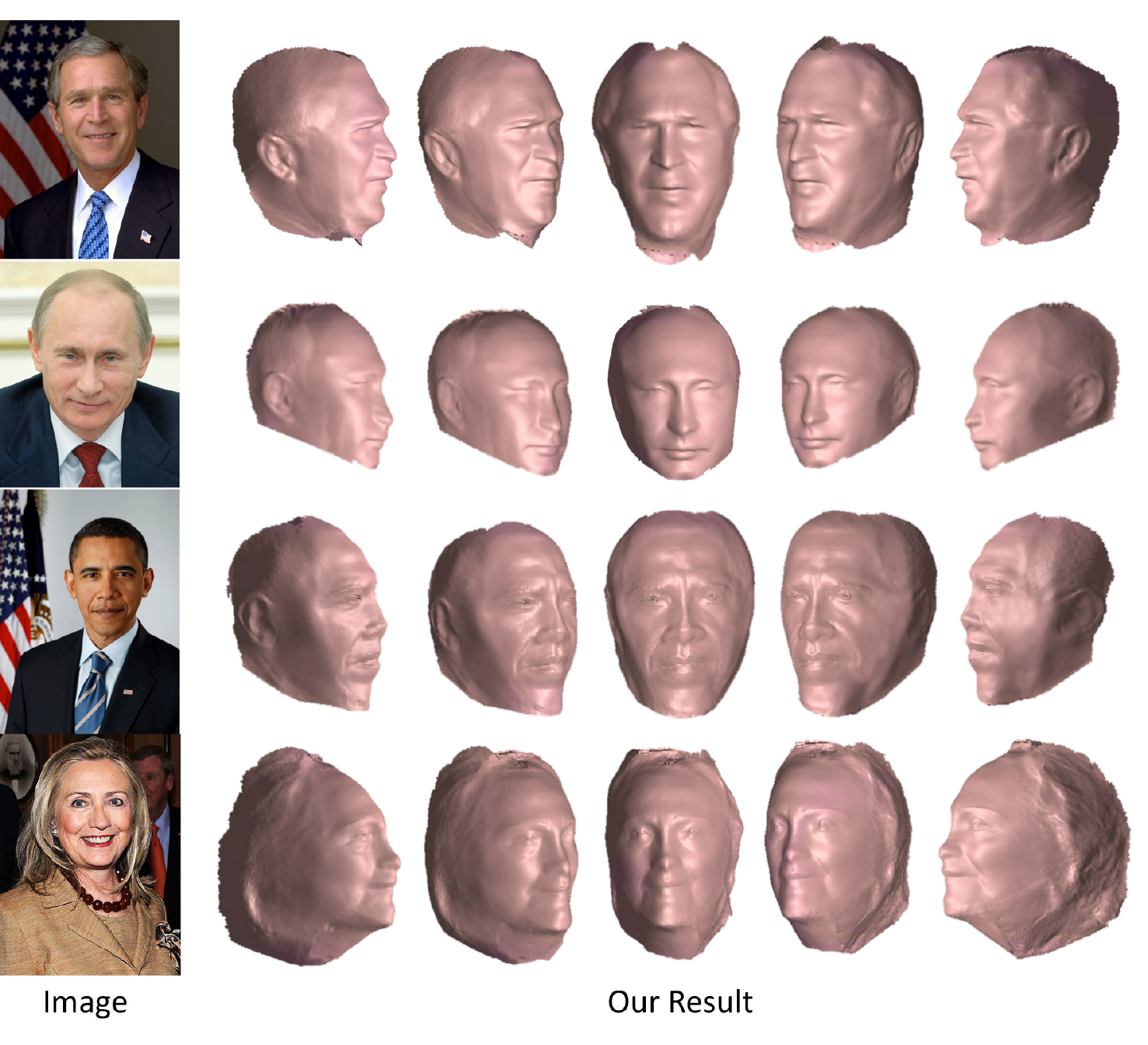}
 	\caption{ Final reconstructed mesh rotated to 5 views to show the reconstruction from all sides. Each color image is an example image among our around $1,000$ photo collection for each person.} \label{fig:result}
 \end{figure}

Fig.~\ref{fig:result} shows the reconstruction result for 4 subjects, each mesh is rotated to five different views. Note that the back and top part of the head are partly missing due to the lack of photos.

To evaluate how the number of photos affects the reconstruction quality, we took 600 photos for George Bush and estimated pose, lighting, texure for each image. We report the L2 intensity difference between the rendered photos and original photos. We tested our reconstruction method with $1/2$, $1/4$, $1/8$ and $1/16$ of the photos in each view cluster (see number of photos per cluster in Table \ref{table:Nphotos}.) The method did not work in $1/16$ case because some view clusters have less than $10$ photos and there was not enough lighting variation within the collection for photometric stereo. Generally, we suggest using more than $100$ photos for frontal view. The number of photos in side view clusters can be smaller (but larger than $30$) because the side view of a human's head is more rigid than the frontal view.

\begin{table}
\begin{center}
\caption{Reconstruction Quality vs. Number of Photos}
\label{table:quality photos}
\begin{tabular}{cccccc}
\hline\noalign{\smallskip}
Number of photos    &N   &N/2   &N/4   &N/8   &N/16\\
\noalign{\smallskip}
\hline
\noalign{\smallskip}
Reprojection Error(intensity)  &$18.29 \pm 4.07$  &$18.70 \pm 4.07$ &$18.71 \pm 4.07$ &$18.80 \pm 4.04$ &$N/A$\\
\hline
\end{tabular}
\end{center}
\end{table}

\begin{figure}
 	\includegraphics[width=\textwidth]{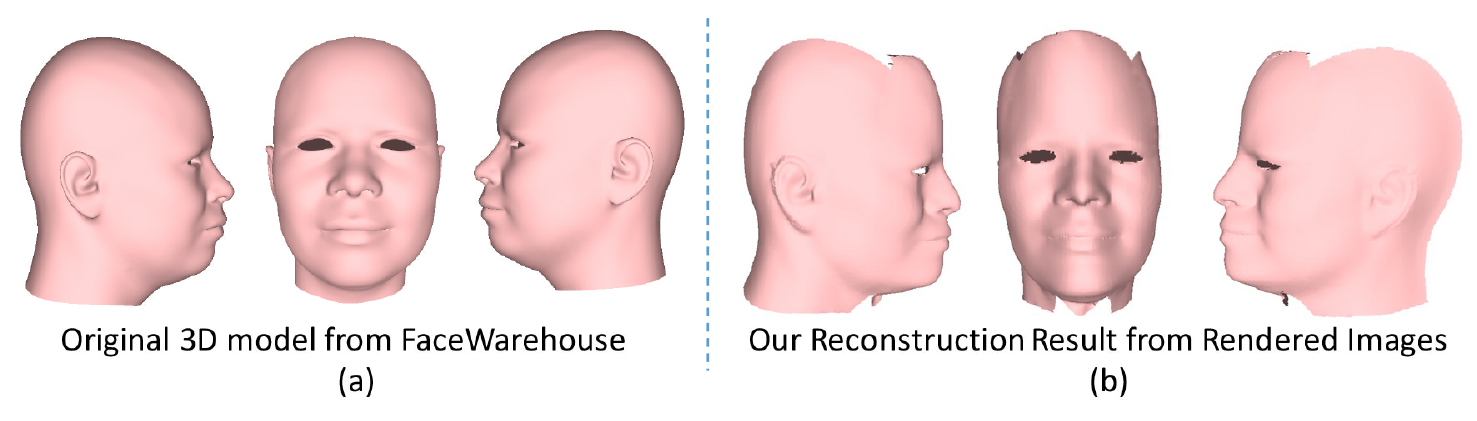}
 	\caption{Reconstruction result from the synthetic photos rendered from a 3D model in FaceWarehouse. The left three shapes are the $-90, 0, 90$ views for the groundtruth shape, and the right three shapes are our reconstruction result.}\label{fig:facewarehouse}
 \end{figure}

We also rendered a 3D model from the FaceWareHouse dataset \cite{cao2014facewarehouse} with 100 lights and 7 poses. We applied our method on these synthetic photos and got a reconstruction result as shown in Fig ~\ref{fig:facewarehouse}. Since we use a template 3D model to correct GBR ambiguity, we cannot get the exact scale of the groundtruth. We do not claim that we have recovered the perfect shape, but the result looks reasonable with an average reprojection error of $11.1 \pm 5.72$.\\

\subsection{Comparison}
\begin{figure}
 	\includegraphics[width=\textwidth]{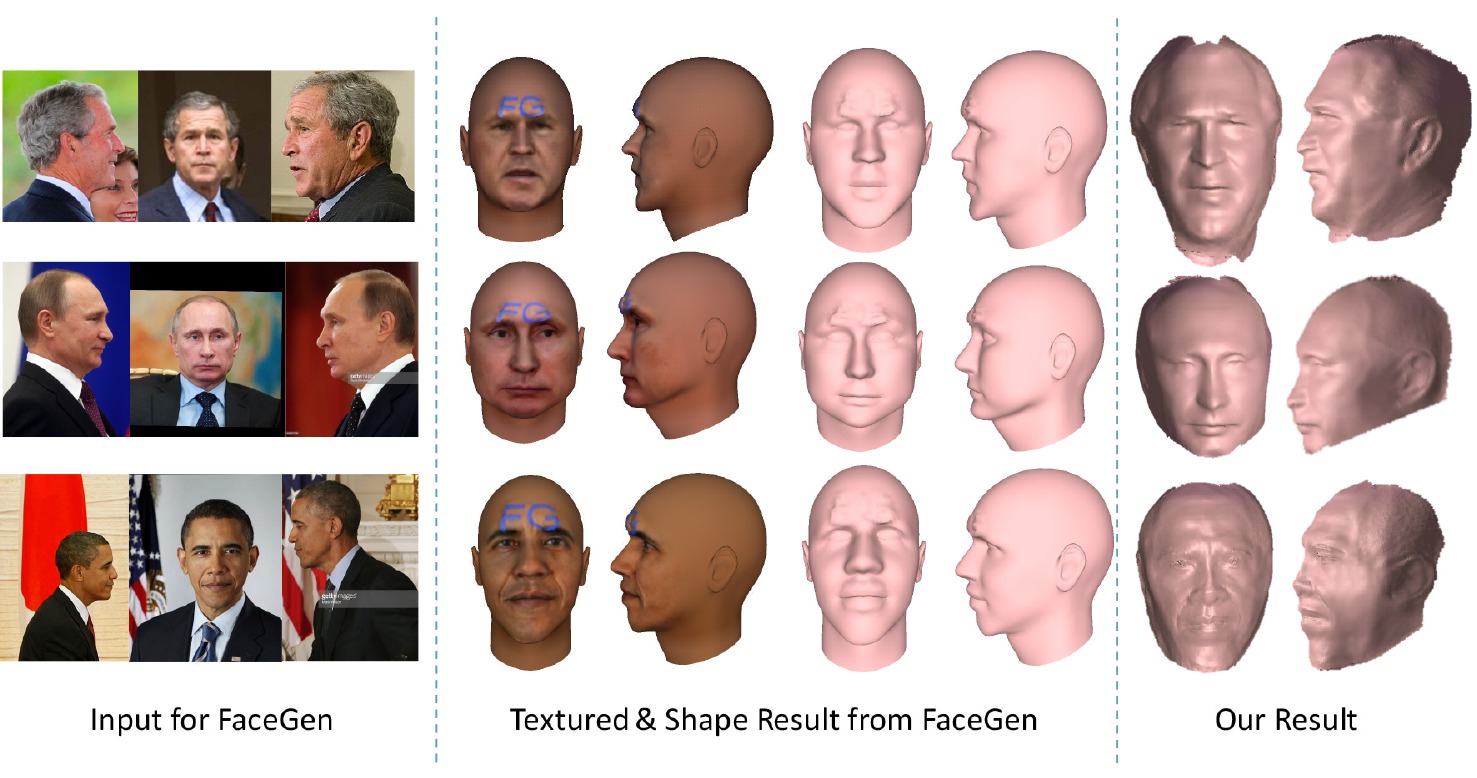}
 	\caption{Comparison to FaceGen (morphable model). We show the textured results and shape results from FaceGen in the middle and our results are on the right as comparisons. Note that the head shape reconstructed by morphable models is average like and not personalized. Additionally, texture hides shape imperfections. }\label{fig:comparison}
 \end{figure}

In Figure~\ref{fig:comparison} we show a comparison to the software FaceGen that implements a morphable model approach. For each person, we manually selected three photos (one frontal view and two side view photos) and used them as the input for FaceGen. The results of FaceGen are too averaged out and not personalized. Note that their ears look the same as each other.

\begin{figure}
 	\includegraphics[width=\textwidth]{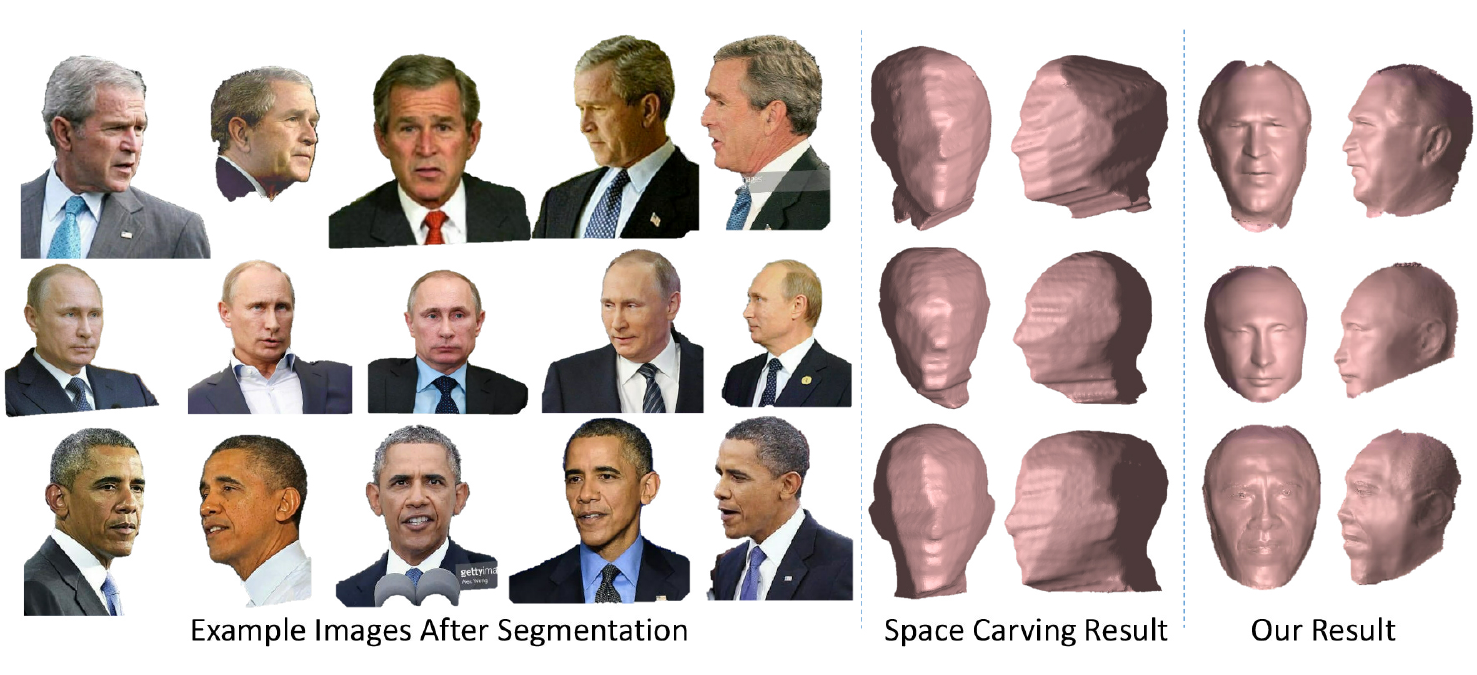}
 	\caption{Comparison to Space Carving method. 5 example segmented images are shown on the left for each person. The segmentations were used as silhouettes. We used around $30$ photos per person.}\label{fig:comparison2}
 \end{figure}

We also tried the Space Carving method \cite{kutulakos2000theory}. For each subject, we manually selected about 30 photos in different poses with a neutral expression. We used the segmentation result obtained from Section~\ref{sec:initial} as the silhouette. We assumed the camera focus length to be 100 and estimated the camera extrinisic parameters using a template 3D head model. We smoothed the carved results using \cite{desbrun1999implicit} and showed the reconstruction in Figure~\ref{fig:comparison2}. The Space Carving method can produce a rough shape of the head. Increasing the number of photos to use does not improve the result.

We have also experimented with VisualSfM \cite{wu2011visualsfm}, but the software could not find enough feature points to run a structure from motion method. This is probably due to the lighting variation and expression change in the photo collection. Similarly, we have tried {\small http://www.123dapp.com/catch}, and it was not able to reconstruct from such photos. 

\begin{table}
\begin{center}
\caption{Reprojection error from 3 reconstruction methods.}
\label{table:compare}
\begin{tabular}{cccc}
\hline\noalign{\smallskip}
Reprojection error    &FaceGen   &Shape-from-Silhouette   &Our method\\
\noalign{\smallskip}
\hline
\noalign{\smallskip}
Bush  &$20.6 \pm 3.80$  &$19.6 \pm 3.55$  &$18.3 \pm 4.04$\\
Putin &$20.1 \pm 4.84$  &$17.2 \pm 4.68$  &$15.1 \pm 5.06$\\
Obama &$21.5 \pm 4.62$ &$20.7 \pm 4.58$  &$19.7 \pm 4.40$\\ 
\hline
\end{tabular}
\end{center}
\end{table}

\begin{figure}
 	\includegraphics[width=\textwidth]{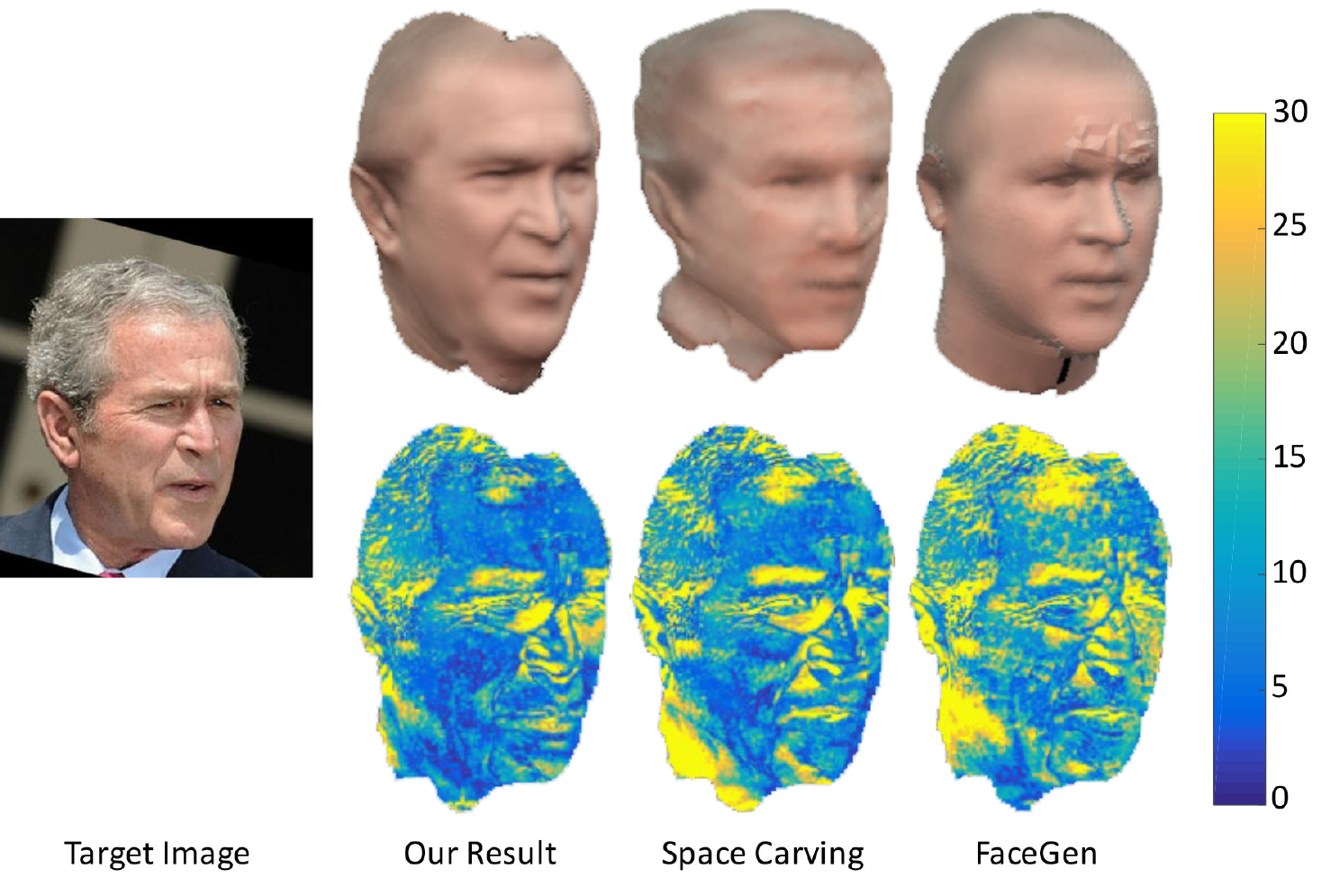}
 	\caption{Visualization of the reprojection error for 3 methods.}\label{fig:errormap}
 \end{figure}

For a quantitative comparison, for each person, we calculated the reprojection error of the shapes from three methods (ours, Space Carving and FaceGen) to $600$ photos in different poses and lighting variations. The 3D shape comes from each reconstruction method. The albedo all comes from average shapes of our clusters, since the Space Carving method and the FaceGen results do not include albedos. The average reprojection error is shown in Table~\ref{table:compare}. The error map of an example image is shown in Fig ~\ref{fig:errormap}. We calculated the error for the overlapping pixels of the three rendered images. Notice that the shapes from FaceGen and Space Carving might look good from the frontal view, but they are not correct when rotating to the target view. See how different the ear part is in the figure.

In future work, we will test the algorithm on more people. Collecting side view images is time consuming. Currently, there are no sets of Internet photos with their corresponding 3D models, thus it is challenging to evaluate quantitatively. We would like to help to solve that by providing our dataset. Furthermore, our GBR ambiguity is just roughly solved by a template model, so the scale might be not exactly the same as the actual mesh. We do not claim to have recovered the perfect shape, but rather show that it is possible to do so from Internet photos, and to encourage further research.


\section{Discussion}

We have shown the first results of head reconstructions from Internet photos. Our method has a number of limitations.  First, we assume a Lambertian model for surface reflectance. While this works well, accounting for specularities should improve results. Second, fiducials for side views were labeled manually; we hope that this application will encourage researchers to solve the challenge of side view fiducial detection. Third, we have not reconstructed a complete model;  the top of the head is missing. To solve this we would need to add photos with different elevation angles, rather than just focusing on the azimuth change. 

We  see several possible extensions to our method. The two we are most excited about are 1) reconstructing 3D non-rigid  motion that includes the head part (not only face, as was done until now), and 2) combining with hair growing methods that can use our reconstructed shape as initialization, e.g., in \cite{chai2015high} the template was produced manually.

\bibliographystyle{splncs03}
\bibliography{eccv2016submission}
\end{document}



\newcommand{\xxx}[1]{\textcolor{blue}{{XXX: #1}}}
\newcommand{\red}[1]{\textcolor{red}{{#1}}}

\pagestyle{headings}
\mainmatter
\def\ECCV16SubNumber{351}  

\title{Head Reconstruction from Internet Photos\\
	Supplementary Material} 

\titlerunning{ECCV-16 submission ID \ECCV16SubNumber}

\authorrunning{ECCV-16 submission ID \ECCV16SubNumber}

\author{Anonymous ECCV submission}
\institute{Paper ID \ECCV16SubNumber}

\maketitle

\begin{figure}
	\includegraphics[width=\textwidth]{figures/putin}
	\caption{3D head reconstruction of Vladimir Putin.  }
\end{figure}

\begin{figure}
	\centering
	\includegraphics[width=.8\textwidth]{figures/nimgs}
	\caption{Number of images used for each person from each view. }
\end{figure}


\bibliographystyle{splncs}
\bibliography{paper}